\newcommand{\CLASSINPUTtoptextmargin}{50mm}
\newcommand{\CLASSINPUTbottomtextmargin}{50mm}
\newcommand{\CLASSINPUTinnersidemargin}{50mm}
\newcommand{\CLASSINPUToutersidemargin}{50mm}
\documentclass[letterpaper, 10 pt, conference]{ieeeconf}  

\IEEEoverridecommandlockouts                              

\usepackage{graphics} 
\usepackage{epsfig} 
\usepackage{mathptmx} 
\usepackage{times} 
\usepackage{amsmath} 
\usepackage{amssymb}  
\usepackage{hyperref}

\usepackage{url}
\usepackage{multirow}

\title{\LARGE \bf
SemVecNet: Generalizable Vector Map Generation \\ for Arbitrary Sensor Configurations
}

\author{Narayanan Elavathur Ranganatha$^{*1}$, Hengyuan Zhang$^{*1}$, Shashank Venkatramani$^{1}$, \\Jing-Yan Liao$^{1}$ and Henrik I. Christensen$^{2}$
\thanks{*These authors contributed equally to this work.}
\thanks{**This research is supported by Nissan and Qualcomm.}
\thanks{$^{1}$Students of Contextual Robotics Institute,
        University of California San Diego, La Jolla, CA 92093, USA 
        {\tt\small \{nelavathurranganatha, hyzhang, svenkatramani, j3liao\}@ucsd.edu}}%
\thanks{$^{2}$Henrik I. Christensen is with Faculty of the Department of Computer Science and Engineering, University of California San Diego,
        La Jolla, CA 92093, USA
        {\tt\small hichristensen@ucsd.edu}}%
\thanks{\copyright 2024 IEEE.  Personal use of this material is permitted.  Permission from IEEE must be obtained for all other uses, in any current or future media, including reprinting/republishing this material for advertising or promotional purposes, creating new collective works, for resale or redistribution to servers or lists, or reuse of any copyrighted component of this work in other works.}%
}

\begin{document}

\maketitle
\thispagestyle{empty}
\pagestyle{empty}

\begin{abstract}

Vector maps are essential in autonomous driving for tasks like localization and planning, yet their creation and maintenance are notably costly. While recent advances in online vector map generation for autonomous vehicles are promising, current models lack adaptability to different sensor configurations. They tend to overfit to specific sensor poses, leading to decreased performance and higher retraining costs. This limitation hampers their practical use in real-world applications. In response to this challenge, we propose a modular pipeline for vector map generation with improved generalization to sensor configurations. The pipeline leverages probabilistic semantic mapping to generate a bird's-eye-view (BEV) semantic map as an intermediate representation. This intermediate representation is then converted to a vector map using the MapTRv2 decoder. By adopting a BEV semantic map robust to different sensor configurations, our proposed approach significantly improves the generalization performance. We evaluate the model on datasets with sensor configurations not used during training. Our evaluation sets includes larger public datasets, and smaller scale private data collected on our platform. Our model generalizes significantly better than the state-of-the-art methods. The code will be available at \href{https://github.com/AutonomousVehicleLaboratory/SemVecNet}{https://github.com/AutonomousVehicleLaboratory/SemVecNet}.


\end{abstract}


\section{INTRODUCTION}

\begin{figure}
\centering
\includegraphics[width=0.48\textwidth]{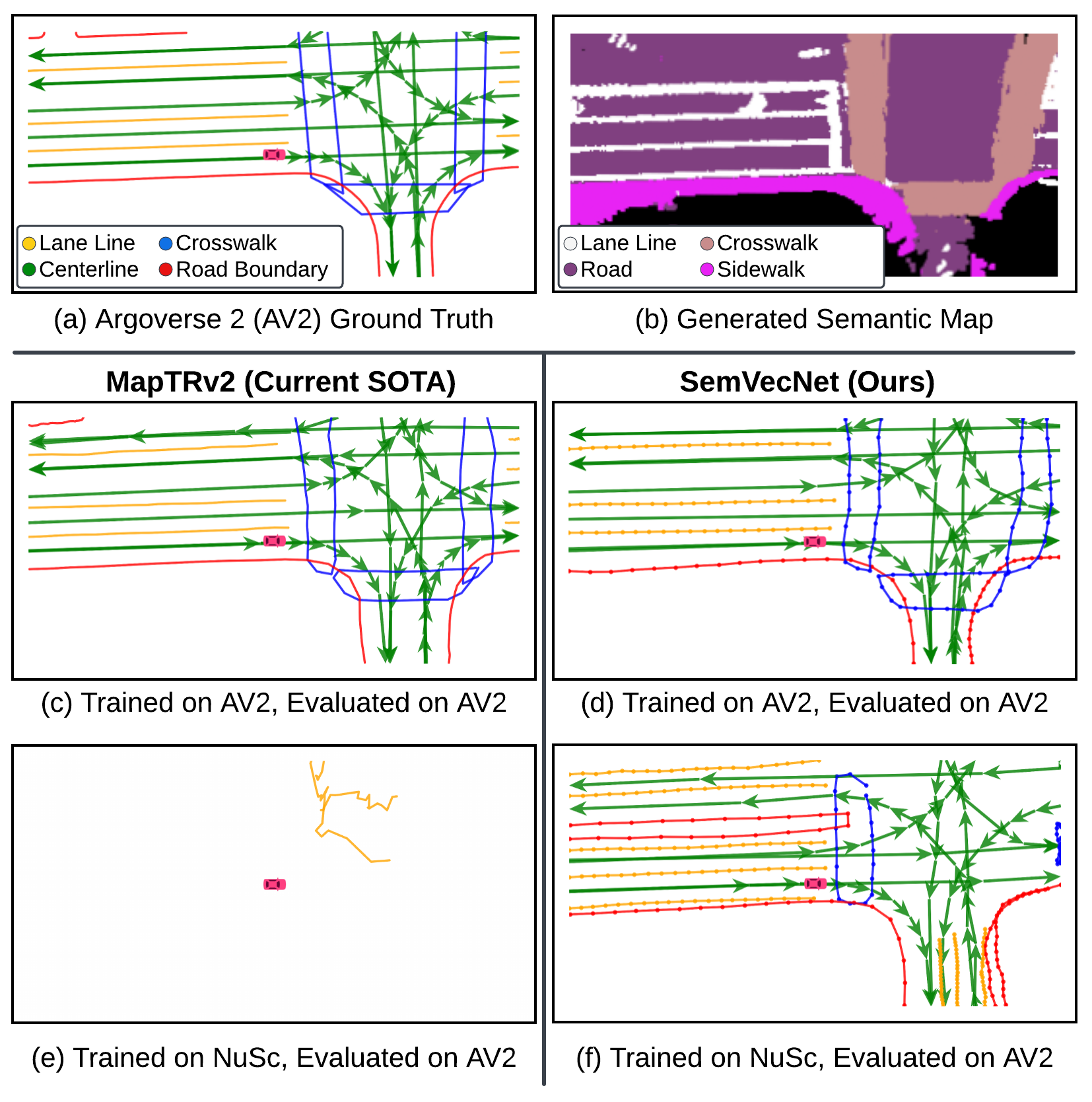}
\caption{State-of-the-art vector map generation models, for example, MapTRv2, perform well when trained and evaluated on the same dataset, but their performance degrades significantly when evaluated on a different dataset. Data labeling and retraining are required which limits their real-world application. SemVecNet shows significant improvements in performance transfer by leveraging the semantic mapping with more robust sensor configuration generalization as an intermediate representation.}
\label{fig:teaser}
\end{figure}

Countless applications await the deployment of autonomous driving technology. For instance, addressing the truck driver shortages in the logistic sector, reducing the need for parking spaces and most importantly, improving driving safety. Current autonomous vehicle technology relies on high-definition (HD) maps. HD maps, as shown in Fig.~\ref{fig:teaser}, consist of accurately localized road features, including lane lines, crosswalks, sidewalks, and centerlines~\cite{caesar_nuscenes_2020}\cite{wang2023openlanev2}\cite{Argoverse2}. These highly detailed maps are essential for localization and planning, and also provide a rich context for detection, tracking, and prediction~\cite{nivash2023simmf_prediction}. For example, a local planner follows the lane labels while avoiding collision with other road users and the crosswalk labels could serve as a strong prior for pedestrian interaction.

Traditionally, HD map creation involves substantial human annotation efforts; however, the dynamic nature of urban environments renders human involvement an ongoing expense. This predicament prompts two potential strategies: a transition to sparse-definition (SD) maps or the automation of HD map generation. While the former option may prove valuable in tasks like localization~\cite{osm_loc_1}\cite{osm_localization} and navigation~\cite{osm_nav_1}\cite{osm_nav_2}, prior research~\cite{osm_av1}\cite{liao2023osm} emphasizes the enduring significance of lane-level information, particularly in complex scenarios like intersections, for trajectory prediction. Some methods propose to automate the HD map generation offline~\cite{Zhou2021IROS_HDmap}. However, with 
about 15,000 miles (24,000 kilometers) increase in urban roads or streets per year in the US from 2011-2021~\cite{road_statistics}, generating offline HD maps is not scalable.

In contrast, contemporary efforts~\cite{Li_hdmapnet}\cite{MapTR}\cite{maptrv2}\cite{li2023toponet}\cite{wu2023topomlp} have pursued directly generating HD maps online, showcasing substantial efficacy within specific sensor configurations. Nevertheless, the primary challenge observed across these approaches is poor generalization to sensor configurations not seen in the training data. As shown in Fig.~\ref{fig:teaser}, when MapTRv2~\cite{maptrv2} is evaluated on a dataset that is different from the training dataset, the model output degrades significantly. The poor generalization impedes the deployment of these models in real-world applications where changing of sensor locations, adding/removing sensors, and deploying on different platforms are common. These changes would incur costly data collection and retraining. Not to mention that this is all based on the assumption that HD map labels are available for retraining. For many including us, the cost is prohibitive.

We identify that the huge performance gap originates from the joint training of BEV view transform module~\cite{philion2020lift}\cite{li2022bevformer} with the decoder. These modules, despite trying to account for sensor configurations by taking camera parameters as input, inevitably overfit to the training set where the sensor configurations are often fixed, resulting in poor generalization on datasets with different sensor configurations. 

Thus, rather than adopting an end-to-end approach with a view transform module, our work presents a modular pipeline designed to address the generalization issue. We leverage a semantic mapping approach that takes a 3D LiDAR point cloud, 2D images, and sensor intrinsics and extrinsics to generate intermediate semantic grid maps with improved generalization to different sensor configurations. The maps are further processed by a vectorization module into vector maps. This method, uses 3D geometry to mitigate overfitting and adapts to varying sensor setups, significantly improving generalization, as shown in Fig.~\ref{fig:teaser}. By standardizing input through this ego-centric BEV map, our model demonstrates enhanced adaptability in diverse real-world scenarios, particularly in vector map generation with unseen sensor platforms, without the need for model retraining.

The paper is organized with an initial discussion of related work in Section~\ref{sec:related-work}. Then we present the overall pipeline in Section~\ref{sec:approach}, followed by the associated experiments and ablation studies in Section~\ref{sec:experiments}. Lastly, we summarize in Section~\ref{sec:conclusion}. Our key contributions are summarized as follows:

\begin{itemize}

\item \textbf{Sensor Configuration Generalizable Vector Map Generation Pipeline: } 

Rather than projecting images into BEV through view transform networks, our approach leverages a BEV semantic map as an intermediate representation to refrain our pipeline from overfitting to any specific sensor configuration. This allows our proposed vector map generation pipeline to generalize to unseen sensor platforms.

\item \textbf{Validation of Generalization Capability through Cross-Dataset and Real-world Experiments: } 

We test our pipeline on datasets with different sensor configurations including real-world data collected on our vehicle platform, which demonstrates that our proposed approach improves performance transfer significantly.

\end{itemize}

\section{Related Work}\label{sec:related-work}

The important role of maps in current autonomous driving architectures motivates extensive research in this area. Some studies focus on understanding scene semantics, which is a crucial step in building maps. Others attempt to directly generate maps with various levels of detail.

\subsection{Semantic Mapping}\label{sec:rel-semantic-mapping}

Prior work explores building a semantic representation of the environment, derived either from a single frame or multiple frames. Single-frame approaches are especially challenging due to the occlusion by other road users and building structures. BEVFormer~\cite{li2022bevformer} addresses this issue by fusing surround-view-camera image features to create a consistent BEV feature for decoding objects and semantic maps.  BEVFusion~\cite{liu2022bevfusion} further enhances this approach by integrating LiDAR point cloud data with image features. These methods demonstrate impressive performance in various downstream tasks, such as detection and mapping. However, their limitation arises from the view transform network overfitting to the training data, making model retraining a necessity when applied to a new setting with different sensor configuration settings, such as a different car with cameras at varied heights and viewing angles.

On the other hand, multi-frame approaches can leverage information extracted over time. Many early works focus on drivable regions~\cite{Sengupta12DenseVisual}\cite{Sengupta13Stereo} and often rely on planar assumptions~\cite{Sengupta12DenseVisual}. More recently Paz et al.~\cite{Paz2020IROSSemanticMapping} proposed the fusion of point cloud maps and image semantic segmentations to generate a probabilistic semantic grid map. The pipeline is more generalizable to sensor configuration as it explicitly takes the camera intrinsics and extrinsics into account in the mapping process. In this study, we adopt the semantic maps as an intermediate representation. We extend the pipeline to utilize real-time point cloud and semantic segmentation to enable online mapping.

While semantic grid maps are informative, they require additional processing to be used in downstream tasks such as route planning and lane following. Additionally, they are less compact than vectorized maps, leading to storage and bandwidth overhead.

\begin{figure*}
\centering
\includegraphics[width=0.9\textwidth]{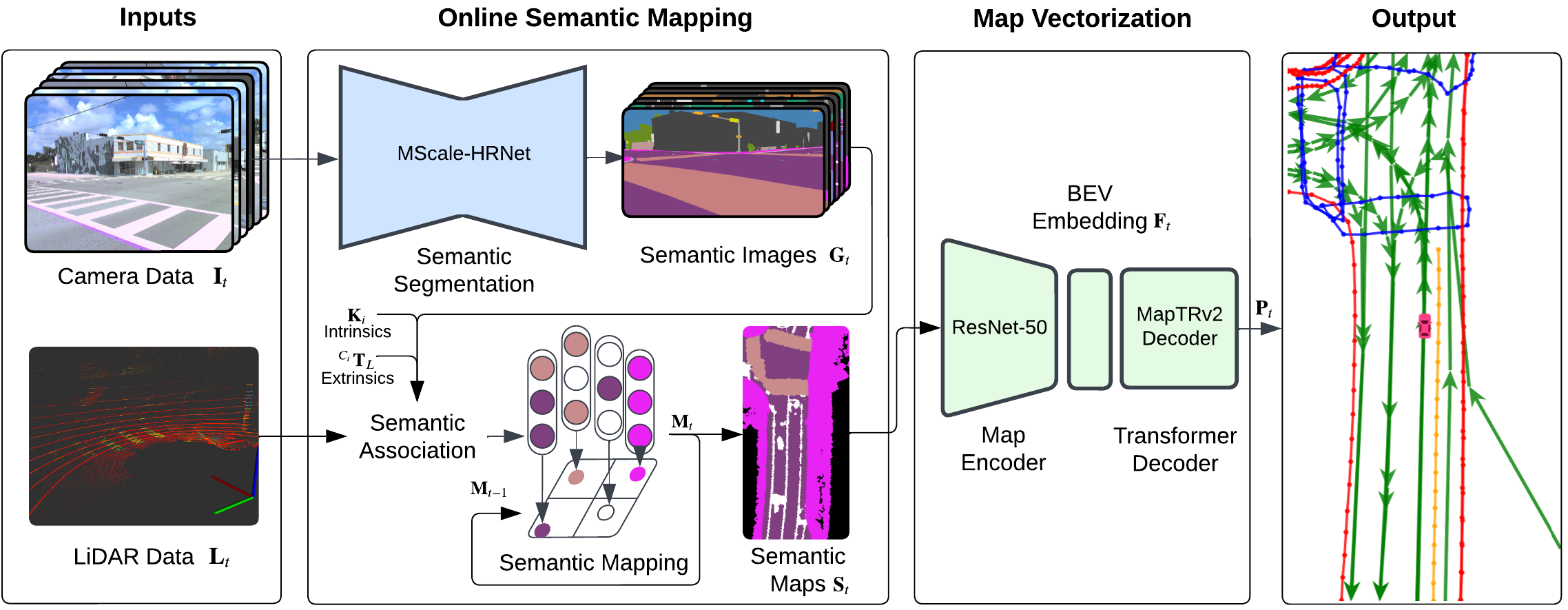}
\caption{SemVecNet takes camera images and LiDAR point cloud to generate a generalized sensor configuration BEV semantic map as an intermediate representation. The semantic map is vectorized into map elements such as centerlines, lane boundaries, crosswalks and road boundaries.}
\label{fig:pipeline}
\end{figure*}

\subsection{Online Vector Mapping}\label{sec:rel-vector}

Vector maps encode map elements as polylines. There are online approaches and offline approaches. Offline approaches allow the aggregation of more information. For example, Zhou et al.~\cite{Zhou2021IROS_HDmap} automate the HD Map building pipeline with instance segmentation, mapping, and particle filter based lane aggregation. However, they require ongoing maintenance.

Recent methods explore online HD map construction to avoid the need for continuous map maintenance. Li et al. propose HDMapNet~\cite{Li_hdmapnet}, which generates a semantic representation with instance features, and then post-processes it to generate vectorized maps. Liu et al. propose VectorMapNet~\cite{liu2022vectormapnet} to decode vector representation directly without the intermediate semantic representation. Liao et al. propose MapTR~\cite{MapTR} and MapTRv2~\cite{maptrv2} which use permutation invariance loss to supervise the vectorized map element generation process. 

A new trend in online mapping not only estimates the map elements but also detects traffic elements and understands relations between them~\cite{wang2023openlanev2}\cite{li2023toponet}\cite{wu2023topomlp}. For example, in the OpenLaneV2~\cite{wang2023openlanev2} dataset, traffic signs in images are associated with the centerlines of the lanes under their control. TopoNet~\cite{li2023toponet} uses a scene graph neural network to model the association. 

In our research, our focus is solely on the map element estimation task. While existing architectures achieve great performance when trained and tested on the same dataset, they often fall short in generalizing to other sensor settings. In contrast, our approach bridges the gap by leveraging a semantic mapping pipeline more robust to different sensor configurations to generate an intermediate semantic representation, and the final map is subsequently derived by our vectorization pipeline.

\section{Approach}\label{sec:approach}

We hypothesize the root cause that prohibits existing work from generalizing across different sensor setups is transformation overfitting. As the network is only trained with one specific sensor configuration in the dataset, they tend to overfit to the coordinate transformation between the sensors and the world. To address the generalization issue in vector map creation approaches, we propose to combine semantic mapping that considers varying intrinsic and extrinsic configurations to generate an intermediate semantic map, and then decode the semantic map into the vector representations.

Formally, the problem is defined as follows: given a set of $n$ images $\mathbf{I}_t = \{I_1, I_2, \dots, I_n\}$, LiDAR point clouds $\mathbf{L}_t$ and semantic map prior $\mathbf{M}_{t-1}$ at timestep $t$, generate a collection of $m$ point sets $\mathbf{P}_t =\{ \mathbf{P}_1, \mathbf{P}_2, \dots, \mathbf{P}_m\}$ to represent road elements such as lane-markings, pedestrian crossing, and road edges. The initial semantic map prior $\mathbf{M}_{0}$ is all zeros and in each timestamp the semantic map is also updated.

We describe the two major components of SemVecNet, semantic mapping and map vectorization in Sub-section~\ref{sec:semantic-mapping} and Sub-section~\ref{sec:map-vectorization}.


\subsection{Real-time Semantic Mapping}\label{sec:semantic-mapping}

We introduce the sensor configuration generalizable semantic mapping to the online vector mapping pipeline. The semantic mapping pipeline generates ego-centric Birds-Eye-View (BEV) semantic map $\mathbf{S}_t \in \mathbb{R}^{H_{bev} \times W_{bev} \times 3}$, where the color channel represents the map types such as road, lane markings, crosswalks and sidewalks. Building on prior work~\cite{Paz2020IROSSemanticMapping}\cite{AVL_semantic_mapping_sensors}, we further make the pipeline real-time and online. It comprises of three key components: semantic segmentation, semantic association, and semantic mapping. 

\textbf{Semantic Segmentation:} The semantic segmentation model takes images $\mathbf{I}_t$ and generates semantic masks $\mathbf{G}_t$. We use HRNet+OCR (MScale-HRNet)~\cite{tao2020hierarchicalMSHRNet} as the model due to its highly accurate segmentations, yielding cleaner semantic maps. We leverage TensorRT~\cite{TensorRT} to optimize MScale-HRNet, which reduces inference time significantly without noticeable performance degradation. This optimization results in the mapping pipeline operating at a frequency exceeding 10 Hz without compromising data quality.

\textbf{Semantic Association:} The semantic association module projects LiDAR point cloud $\mathbf{L}_t$ onto the semantic images $\mathbf{G}_t$ to associate depth with semantic, resulting in a semantic point cloud with accurate geometry. Given the camera intrinsics $\mathbf{K}_i$ and extrinsics ${}^{C_i}\mathbf{T}_{L}$ for the camera $i$, the projection of a point $\mathbf{x}_L$ in LiDAR into a pixel in image $\mathbf{x}_{I}$ is given by

\begin{equation}
    \mathbf{x}_{I} =  \mathbf{K}_i [\mathbf{I}^{3\times3} | \mathbf{0} ] {}^{C_i}\mathbf{T}_{L}\mathbf{x}_L,
\end{equation}
where the extrinsics ${}^{C_i}\mathbf{T}_{L}$ is a transformation from LiDAR frame to Camera frame. Different from prior approach~\cite{Paz2020IROSSemanticMapping}\cite{AVL_semantic_mapping_sensors} where they use LiDAR point map, we use real-time point cloud which eliminates the point cloud map building process, enabling it to be used in online mapping. 

\textbf{Probabilistic Mapping:} The probabilistic mapping module takes the semantic point cloud 
$\mathbf{C}_t$ and integrates it into a BEV probabilistic grid map $\mathbf{M}_t \in \mathbb{R}^{H_{bev} \times W_{bev} \times C_{bev}}$, where $C_{bev}$ is the number of class labels. The grid map is then rendered into a semantic map $\mathbf{S}_t$ with the highest probabiltiy class. 

\begin{equation}
    P(c_t|\mathbf{M}_{t-1}, z_t, i_{t}) = \frac{1}{N_m}P(z_t|c_t)P(i_t|c_t)P(c_{t-1}|\mathbf{M}_{t-1}),
\end{equation}

with $z_t$ and $c_t$ represent the observed and actual semantic label for the grid at time $t$, $i_t$ represents the observed intensity and $N_m$ represents the normalization factor. The confusion matrix of the model $P(z_t|c_t)$ and LiDAR intensity prior $P(i_t|c_t)$ are accounted for in the process.

Combined with suitable SLAM methods~\cite{Campos_orbslam3_2021}\cite{Li2021liliom}, the semantic mapping pipeline can run online at more than 10Hz and build a semantic map with a single camera and LiDAR. Our approach leverages projective geometry and therefore can build generalized sensor configuration semantic representation with different cameras and LiDARs. The semantic map with roads, lanes, crosswalks and sidewalks can then be used for vector mapping.

\subsection{Map Vectorization}\label{sec:map-vectorization}
Given the generated BEV semantic map $\mathbf{S}_t \in \mathbb{R}^{H_{bev} \times W_{bev} \times 3}$, the Map Vectorization Model (MVM) takes $\mathbf{S}_t$ and maps it to vector map elements $\mathbf{P}_t=[\mathbf{P}_1, \mathbf{P}_2, \dots, \mathbf{P}_m]$. Each vector map element is modeled as a point set $\mathbf{P}_j = [p_0, p_1, \dots, p_n]$. We utilize the equivariant permutations as stated in~\cite{maptrv2} to deal with the ambiguity of multiple correct permutations for a $\mathbf{P}_j$.

The MVM has an encoder-decoder format. Given $\mathbf{S}_t$, a feature map $\mathbf{F}_t \in \mathbb{R}^{H_c\times W_c \times C}$ is generated using a encoder. This encoder in principle can be any model that takes an image as input and outputs a feature representation. We adopt the Resnet-50~\cite{He_2016_resnet} model as our image encoder with reduced strides to maintain the dimension. $\mathbf{F}_t$ acts as BEV embedding which is then passed into the decoder. 

The decoder following a similar structure to the one in MapTRv2~\cite{maptrv2}, consists of a transformer decoder that uses map queries. These queries consist of instance-level queries $Q_i = \{ q_i^j\}_{j=1}^{m}$ as well as point-level queries $Q_p = \{ q_p^j \}_{j=1}^{n}$ that each instance-level query uses. These are then summed to form hierarchical queries, so for map element $k$, we get $Q_h^k = \{q_h^{j}\}_{j=1}^{n} = \{ q_i^k + q_p^j \}_{j=1}^{n}$. The decoder uses self-attention within the map queries to facilitate information exchange between instance-level queries and also the point-level queries within the instances. The decoder then uses cross-attention to facilitate the interaction between the map queries and $\mathbf{F}_t$. 

The output of the decoder is then passed into a classification branch which outputs the instance scores and a point regression branch which produces the 2D coordinates of the $n$ points in the point set. This way each query $Q_h^{k}$ is mapped to a class $c_{vk}$ = \{ Pedestrian Crossing, Lane Divider, Lane Boundary, Centerline\} and each hierarchical query $Q_h^{kj}$ is mapped to a 2D BEV point $p_{kj} = \{ x_{kj}, y_{kj} \}$. 

\textbf{Loss:} To supervise the MVM, following MapTRv2~\cite{maptrv2}, we perform instance-level matching and then point-level matching. Let $\hat{\sigma}_k$ denote the optimal matching and ordering for the $k^{th}$ map element in the ground truth. Therefore $\hat{c}_{v\hat{\sigma}_k}$ denotes optimal instance matching to ground truth element $k$ and $\hat{P}_{\hat{\sigma}_k} = \{\hat{p}^j_{\hat{\sigma}_k}\}_{j=1}^m$ represents optimal ordering of the points once instance matching is done. $\hat{e}_{\hat{\sigma}_k}^j$ denotes the predicted edge between the points $\hat{p}^j_{\hat{\sigma}_k}$ and $\hat{p}^{(j+1)\, mod \, n}_{\hat{\sigma}_k}$ and $e_{k}^j$ denotes the predicted edge between the points $p^j_{k}$ and $p^{(j+1)\, mod \, n}_{k}$. We follow~\cite{maptrv2} and use the focal loss $l_{focal\_loss}$~\cite{Lin_2020_focal_loss} for the instance label classification $l_{cls}$, point-to-point loss $l_{p^2p}$ for each predicted point and an edge direction loss $l_{dir}$ to supervise the direction of the edge connecting two points. We also use the auxillary BEV segmentation loss $l_{Seg}$ introduced in MapTRv2. For this a auxillary segmentation head is added which is denoted as $\mathbf{s}(\cdot)$. We do not use the auxillary perspective view segmentation and depth estimation losses mentioned in MapTRv2 as the perspective view is not part of the input. The losses are defined as follows:

\begin{eqnarray}
    l_{cls} = \sum_{k=1}^{m} l_{focal\_loss}(\hat{c}_{v\hat{\sigma}_k},  c_{vk})\\
    l_{p^2p} = \sum_{k=1}^{m} \mathbf{1}_{c_k \ne \phi} \sum_{j=1}^{n} D(\hat{p}^j_{\hat{\sigma}_k}, p_k^j)\\
    l_{dir} = \sum_{k=1}^{m} \mathbf{1}_{c_k \ne \phi} \sum_{j=1}^{n} cos(\hat{e}_{k}^j, e_{k}^j)\\
    l_{Seg} = l_{CE}(\mathbf{s}(\mathbf{F}_t), \mathbf{GT}_{BEV})
\end{eqnarray}

where $D()$ denotes the Manhattan distance and $cos()$ denotes cosine similarity. $l_{CE}$ is the cross-entropy loss and $\mathbf{GT}_{BEV}$ denotes the ground truth BEV Segmentation. Therefore the final loss $l$ becomes:
\begin{equation}
    l = w_{cls}l_{cls} + w_{p^2p}l_{p^2p} + w_{dir}l_{dir} + w_{Seg}l_{Seg}
\end{equation}

\section{Experiments}\label{sec:experiments}

In this section, we perform experiments to validate the effectiveness of SemVecNet. We introduce the datasets, evaluation metrics and training configurations in Sub-section~\ref{sec:exp-dataset-metric}. Then we perform experiments to show the generalization performance with cross-dataset experiments in Sub-section~\ref{sec:exp-cross-dataset} and real-world data in Sub-section~\ref{sec:exp-real-world}. Additionally we present various ablation studies to show the key factors that affect the design in Sub-section~\ref{sec:exp-ablations}. Lastly we discuss limitations and potential improvements in Sub-section~\ref{sec:discussion-limitation}.

\subsection{Datasets, Metric and Training}\label{sec:exp-dataset-metric}

\textbf{Datasets:} We conducted evaluations of our method using two prominent datasets: Argoverse2~\cite{Argoverse2} and the NuScenes~\cite{caesar_nuscenes_2020} dataset. Argoverse2 offers 3D Vector Maps for each 15-second log, featuring images sampled at 20Hz and LiDAR data sampled at 10Hz. The dataset comprises 700 training logs, 150 validation logs, and 150 test logs. In our approach with Argoverse2, we operate at the frequency of the LiDAR sensor, utilizing data sampled at 10Hz and also only do 2D Vector Map estimation. On the other hand, NuScenes provides a collection of 1000 scenes, accompanied by city-level Vector Maps. In the case of NuScenes, we specifically leverage the samples dataset as it contains synchronised data, operating on sensor data sampled at 2Hz to train our model.

\begin{figure}
\centering
\includegraphics[width=0.35\textwidth]{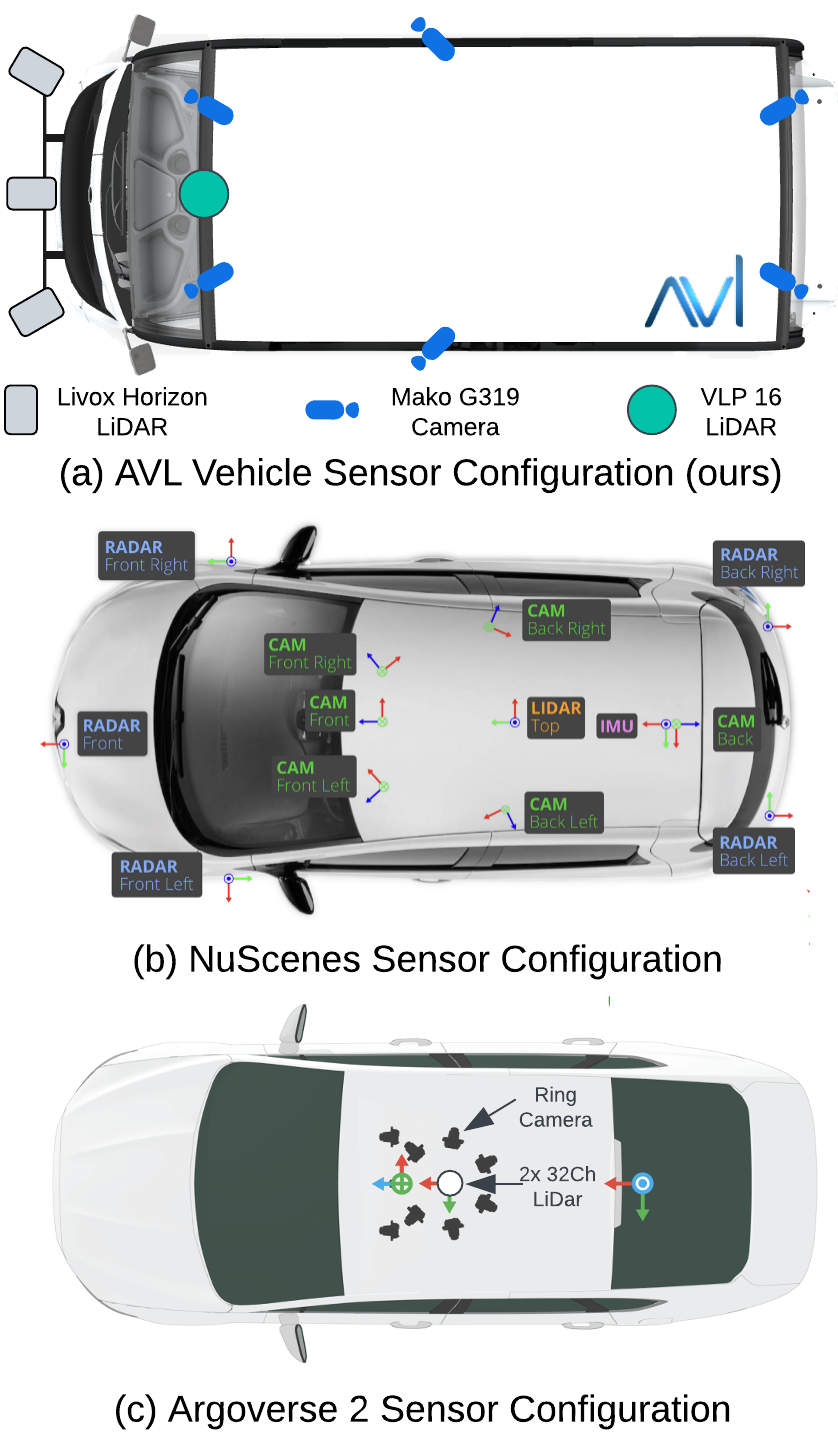}
\caption{The diagrams (a), (b), and (c) display the significant configuration changes across platforms in sensor poses and number of sensors, for both LiDAR and cameras in NuScenes~\cite{caesar_nuscenes_2020}, Argoverse 2~\cite{Argoverse2} and AVL.}
\label{fig:avl_cart}
\end{figure}

\textbf{Metrics:} We follow MapTRv2~\cite{maptrv2} and use Average Precision (AP) metric for evaluation. The AP is calculated for \textit{Pedestrian Crossing}, \textit{Lane Divider}, \textit{Road Boundary} and \textit{Centerlines} separately and averaged. The AP is calculated with chamfer distance $D_{Chamfer}$ across three thresholds $\boldsymbol{\theta} \in \{0.5m, 1.0m, 1.5m\}$. The Chamfer distance is given by  
\begin{equation}
    D_{Chamfer}(X,Y) = \frac{1}{|X|}\sum_{x \in X} \min_{y \in Y} || x - y ||_2^2 + \frac{1}{|Y|}\sum_{y \in Y} \min_{x \in X} || y - x ||_2^2.
\end{equation}
The AP for each class is averaged to give the Mean Average Precision (mAP), given by
\begin{equation}
    AP = \frac{1}{|\boldsymbol{\theta}|} \sum_{\lambda \in \boldsymbol{\theta}} AP_{\lambda}.
\end{equation}

\textbf{Training:} For the Semantic Segmentation Network, we use HRNet-OCR trained on the Mapillary dataset~\cite{MVD2017_Mapillary_Vistas} as described in~\cite{AVL_semantic_mapping_sensors}. For the decoder, we remove the perspective view to BEV view transfomation from the baseline MapTRv2 model and pass encoded BEV features to the MapTRv2 Decoder. We follow the same learning rate scheduler as MapTRv2 but utilize cyclic momentum as we observed that led to better convergence. Loss weghts are $w_{cls}=2.0, w_{p2p}=5.0, w_{dir}=0.005, w_{Seg}=1.0$ following~\cite{maptrv2}. The model takes 12 hours to converge on 8 NVIDIA A-10 GPUs for both datasets with a batch size of 5 on each GPU. 

\subsection{Cross-dataset Experiments}\label{sec:exp-cross-dataset}

We are particularly interested in quantitatively measuring the model's performance generalization capabilities. This is one of the most critical issues faced in real-world applications. A model with great performance only on its trained sensor configuration has a prohibitive cost to adapt to new platforms and will not be scalable in the long run. 

To this end, we conduct cross-dataset experiments and use the mAP ratio to measure the performance transfer. In cross-dataset experiments, a model trained on dataset $A$ is evaluated on the validation set of dataset $A$ and $B$. The ratio of the mAP on the test set of $B$ to $A$ measures performance transfer.

This task is especially challenging since the cross-dataset sensor configurations are often drastically different. For example, NuScenes dataset has six cameras and a LiDAR and Argoverse 2 has seven cameras and two LiDARs, as shown in Fig.~\ref{fig:avl_cart}. These sensors are also placed differently, resulting in varying viewing angles. This explains why MapTRv2, the current SOTA for vector map generation, achieves an mAP ratio of 0, as shown in Table~\ref{tab:cross-dataset}. Comparatively, SemVecNet obtains a performance transfer of 24.8\% when evaluating an AV2 trained model on NuScenes, and a 33.1\% performance transfer when evaluating a NuScenes model on AV2. 

From our cross-dataset results, it is apparent that the standard MapTRv2 formulation overfits to the camera configuration of the training setup, limited by learned view transform modules. Alternatively, SemVecNet has a more standardized intermediate representation of semantic maps across datasets, improving performance transfer.


\begin{table}[htbp]
\centering
\caption{Cross-Dataset Performance Transfer}
\begin{tabular}{|c||c|c|c|c|}
\hline
Model    & Train & Test & mAP $\uparrow$ & mAP ratio (\%) $\uparrow$  \\ \hline \hline
\multirow{2}{*}{SemVecNet}  & AV2       & NuSc & 12.2          & \multirow{2}{*}{\textbf{24.8}} \\ \cline{2-4}
         & AV2       & AV2      & 49.0          &  \\ \hline
\multirow{2}{*}{MapTRv2}  & AV2       & NuSc & 0           & \multirow{2}{*}{0} \\ \cline{2-4}
         & AV2       & AV2      & \textbf{67.4}           &  \\ \hline \hline
\multirow{2}{*}{SemVecNet}  & NuSc  & AV2      & 16.2           & \multirow{2}{*}{\textbf{33.1}} \\ \cline{2-4}
         & NuSc  & NuSc & 48.8           &  \\ \hline
\multirow{2}{*}{MapTRv2}  & NuSc  & AV2      & 0           & \multirow{2}{*}{0} \\ \cline{2-4}
         & NuSc  & NuSc & \textbf{61.5}           &  \\ \hline
\end{tabular}
\label{tab:cross-dataset}
\end{table}

\subsection{Real-World Experiments}\label{sec:exp-real-world}

Aside from demonstrating our systems generalization capabilities between datasets, we also conducted experiments with data we collected on UC San Diego campus. The sensor configuration is shown in Fig.~\ref{fig:avl_cart}, and other details of our autonomous driving platform can be found in ~\cite{christensen21:avl}. 

\begin{figure}
\centering
\includegraphics[width=0.3\textwidth, angle=90]{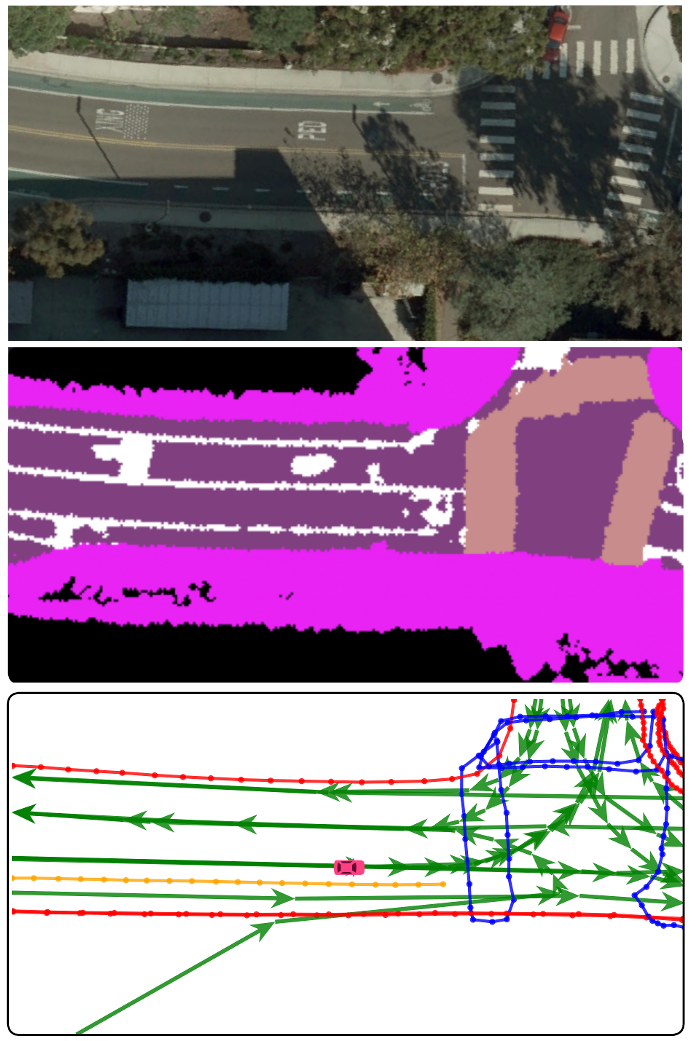}
\caption{The qualitative result by directly inference from campus data. The images from top to bottom are satellite image~\cite{esri}, BEV semantic map, and vector map output from SemVecNet from the same region on UC San Diego campus.}
\label{fig:real-world}
\end{figure}

Given that it is costly to generate groundtruth HD map labels, we only show qualitative results on the campus data to complement our cross-dataset quantitative evaluation. From Fig.~\ref{fig:real-world}, we can see that our pipeline is still able to generate meaningful output without fine-tuning the model with our data. The centerlines and road boundaries are captured accurately for the most part. This experiment shows that our pipeline can work with sensor configurations that are vastly different from the training domain.

\subsection{Ablation Studies}\label{sec:exp-ablations}

We perform experiments to validate various design choices such as single view vs surround view and map resolution. These experiments are performed on Argoverse 2 dataset~\cite{Argoverse2}.

\textbf{Single View vs Surround View:} Our pipeline can adapt to single-camera or multi-camera input. For single-camera input, we take the front camera images and crop them to maintain a similar aspect ratio as a typical semantic segmentation input. For surround-view input, the semantic point cloud incorporates semantic labels from all views to generate the probabilistic semantic map.

\begin{figure}
\centering
\includegraphics[width=0.48\textwidth]{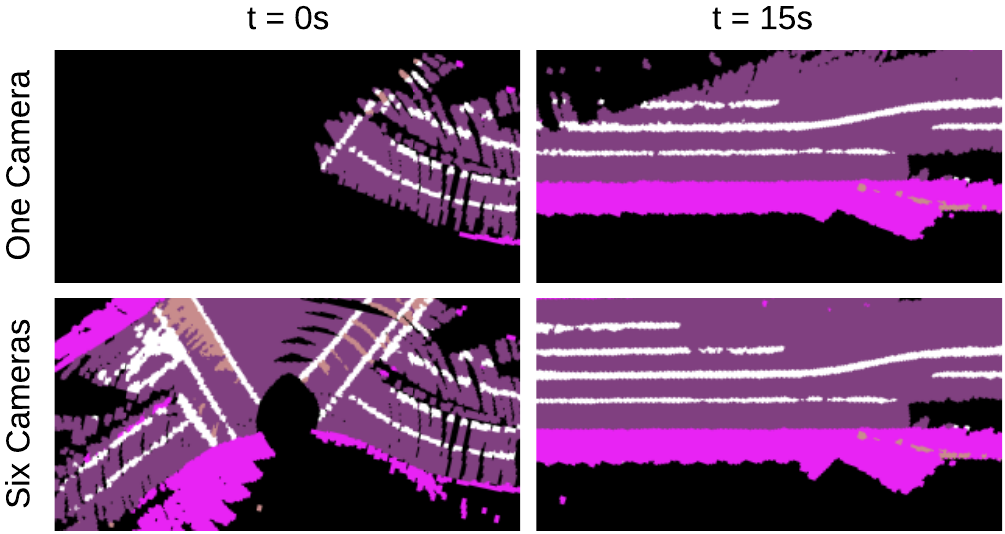}
\caption{Top row represents semantic map made with a single camera. The bottom row represents semantic maps made with six cameras. At the start of a log (first column), a lot of information is lost if some cameras are left out. The map by the end of the log (second column) ends up looking similar.}
\label{fig:ablation-camera-number}
\end{figure}

Compared to maps generated from single-view images, the maps generated from surround-view images have more coverage. Better coverage provides more context, especially for the initial few frames in each scene, as shown in Fig.~\ref{fig:ablation-camera-number}. The lower coverage causes a performance decrease of 28\%.

\begin{table}[htbp]
\centering
\caption{Effect of Number of Cameras}
\begin{tabular}{|c||c|c|}
\hline
 & Number of Cameras & mAP $\uparrow$ \\ \hline \hline
\multirow{2}{*}{Argoverse 2} & 1 & 35.2 \\ \cline{2-3} 
                                   &       6         &       49.0        \\ \hline
\end{tabular}
\label{tab:ablation-camera-number}
\end{table}

We find it interesting that our approach can maintain this level of performance despite reducing the camera number to 1. The temporal aggregation baked into the semantic mapping pipeline allows us to maintain a large portion of our performance even if we reduce the number of cameras.

\textbf{Effect of Semantic Map Resolution:} As mentioned in \ref{sec:semantic-mapping}, grid resolution for the semantic map is a hyperparameter that influences this map creation process. In~\cite{Paz2020IROSSemanticMapping} the grid pixel size is 0.2 meters, however we observe feature blurring in the semantic maps. To try and mitigate this we reduced the pixel size to 0.1 meters. As shown in Table~\ref{tab:res-ablation}, the higher grid resolution doesn't improve performance while significantly increasing the computation. The lack of performance improvement could be due to reduced probabilistic correction, or the network not benefiting sharper images. Based on the results, we use 0.2 resolution for other experiments.


\begin{table}[htbp]
\centering
\begin{tabular}{|c||c|c|}
\hline
 & Pixel Size & mAP $\uparrow$ \\ \hline \hline
\multirow{2}{*}{Argoverse 2} & 0.1 & 48.9 \\ \cline{2-3} 
                                   &       0.2         &       49.0        \\ \hline
\end{tabular}
\caption{Resolution Performance}
\label{tab:res-ablation}
\end{table}

\begin{figure}
\centering
\includegraphics[width=0.48\textwidth]{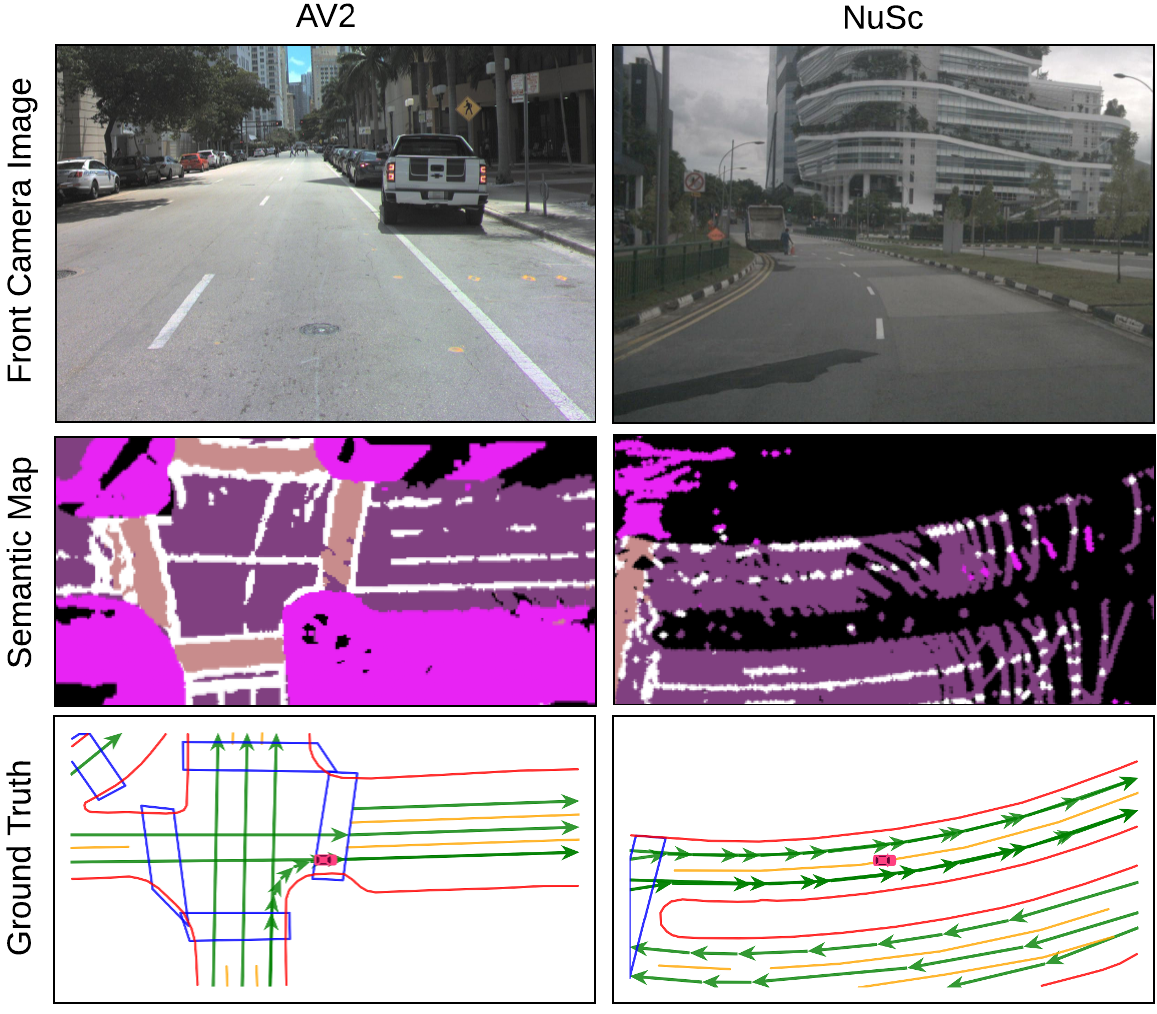}
\caption{This figure shows a significant distribution shift between AV2 and NuScenes. NuScenes semantic maps are sparser due to sparser LiDAR observations. Additionally, NuScenes includes left-hand drive in Singapore logs, while Argoverse 2 is solely U.S. right-hand drive.}
\label{fig:dataset-compare}
\end{figure}

\subsection{Discussion and Limitations}\label{sec:discussion-limitation}

Our proposed approach addresses the critical issue of designing online vector mapping pipelines that are more generalizable improving cross-dataset performance significantly, thus enabling the model to be applied to a new domain without costly labeling and retraining. While our model is designed to be more robust to different camera intrinsics and extrinsics, the performance is still affected by variations in data distribution. 

The first variation is the density of LiDAR observations. AV2 has two stacked 32 channel LiDARs (64 channels effective) while NuScenes has a single 32 channel LiDAR. This causes sparser semantic maps for NuScenes, as seen in Fig.~\ref{fig:dataset-compare}. These semantic map holes presents a domain gap for inputs to the Vector Map generation, and likely hinders cross dataset performance.

Additionally, AV2 is recorded in U.S. cities while NuScenes is recorded both in the U.S. and Singapore. This presents a secondary domain gap in road network distribution. For example, AV2 typically includes a crosswalk at nearly every intersection, a pattern not consistently observed in NuScenes. Consequently, the model trained on AV2 data tends to over-predict the presence of pedestrian crossings at intersections when applied to NuScenes data, leading to reduced accuracy in identifying pedestrian crossings as seen in Table~\ref{tab:individual-class}. The NuScenes logs from Singapore also include left-hand driving scenarios, which is fundamentally different from Argoverse 2's right-hand driving, leading to opposite centerline directions in cross-dataset visualizations. 

\begin{table}[htbp]
\centering
\begin{tabular}{|c|c||cccc|c|}
\hline
\multirow{2}{*}{Train} & \multirow{2}{*}{Test} &       &       & AP $\uparrow$ &       &\multirow{2}{*}{mAP $\uparrow$}\\
          &          & ped.  & div.  & bou.           & cent. & \\
\hline \hline
AV2       & NuSc     & 5.0 & 8.7  & 15.7         & 19.4 & 12.2\\
\hline
AV2       & AV2      & 40.7& 51.3 & 53.4        & 50.6 & 49.0\\  
\hline \hline
NuSc  & AV2          & 3.5& 13.1  & 27.8         & 20.2 & 16.2\\
\hline
NuSc  & NuSc         & 41.4& 50.5 & 54.5         & 49.0 & 48.8\\  
\hline
\end{tabular}
\caption{Individual Class Performance Transfer}
\label{tab:individual-class}
\end{table}


We believe the sparse semantic maps and presence of more diverse road structures in NuScenes likely leads to better generalization and higher performance of NuScenes models on AV2, than AV2 models on NuScenes. 

However, our model shows inferior performance when evaluated on the same dataset. This can be caused by the information loss introduced in the intermediate map representation. Future directions involve using probabilistic grids or neural features as the intermediate representation to reduce information loss and improve performance.

\section{CONCLUSIONS}\label{sec:conclusion}

We introduce SemVecNet, an online vector map generation pipeline designed to be more robust to sensor configurations. Through cross-dataset evaluations, we show that SemVecNet significantly improved the generalization capability, supported by both quantitative and qualitative assessments. Furthermore, SemVecNet showcases its transferability through successful testing on real-world data collected from the UC San Diego campus. We believe this is a step towards sensor-configuration-agnostic autonomous system design that is more scalable, enabling algorithms to be deployed in vehicles with various sensor configurations. Future work involves characterizing the generalization problem arising from view-transform modules in other tasks such as detection and semantic segmentation, and reduce the gap in the proposed pipeline.

\addtolength{\textheight}{-10.5cm}   









\bibliographystyle{unsrt}
\bibliography{ref}

\begin{thebibliography}{10}

\bibitem{caesar_nuscenes_2020}
Holger Caesar, Varun Bankiti, Alex~H. Lang, Sourabh Vora, Venice~Erin Liong, Qiang Xu, Anush Krishnan, Yu~Pan, Giancarlo Baldan, and Oscar Beijbom.
\newblock nuscenes: A multimodal dataset for autonomous driving.
\newblock In {\em 2020 IEEE/CVF Conference on Computer Vision and Pattern Recognition (CVPR)}, pages 11618--11628, Seattle, WA, USA, 13-19 June 2020.

\bibitem{wang2023openlanev2}
Huijie Wang, Tianyu Li, Yang Li, Li~Chen, Chonghao Sima, Zhenbo Liu, Yuting Wang, Shengyin Jiang, Peijin Jia, Bangjun Wang, Feng Wen, Hang Xu, Ping Luo, Junchi Yan, Wei Zhang, and Hongyang Li.
\newblock Openlane-v2: A topology reasoning benchmark for unified 3d hd mapping.
\newblock In A.~Oh, T.~Neumann, A.~Globerson, K.~Saenko, M.~Hardt, and S.~Levine, editors, {\em Advances in Neural Information Processing Systems}, volume~36, pages 18873--18884. Curran Associates, Inc., 2023.

\bibitem{Argoverse2}
Benjamin Wilson, William Qi, Tanmay Agarwal, John Lambert, Jagjeet Singh, Siddhesh Khandelwal, Bowen Pan, Ratnesh Kumar, Andrew Hartnett, Jhony~Kaesemodel Pontes, Deva Ramanan, Peter Carr, and James Hays.
\newblock Argoverse 2: Next generation datasets for self-driving perception and forecasting.
\newblock In {\em Neural Information Processing Systems Track on Datasets and Benchmarks (NeurIPS Datasets and Benchmarks 2021)}, online, 06 -- 14 December 2021.

\bibitem{nivash2023simmf_prediction}
Vidyaa~Krishnan Nivash and Ahmed~H. Qureshi.
\newblock Simmf: Semantics-aware interactive multiagent motion forecasting for autonomous vehicle driving.
\newblock {\em arXiv preprint arXiv:2306.14941}, 2023.

\bibitem{osm_loc_1}
Younghun Cho, Giseop Kim, Sangmin Lee, and Jee-Hwan Ryu.
\newblock Openstreetmap-based lidar global localization in urban environment without a prior lidar map.
\newblock {\em IEEE Robotics and Automation Letters}, 7(2):4999--5006, 2022.

\bibitem{osm_localization}
Philipp Ruchti, Bastian Steder, Michael Ruhnke, and Wolfram Burgard.
\newblock Localization on openstreetmap data using a 3d laser scanner.
\newblock In {\em 2015 IEEE international conference on robotics and automation (ICRA)}, pages 5260--5265. IEEE, 2015.

\bibitem{osm_nav_1}
Matthias Hentschel and Bernardo Wagner.
\newblock Autonomous robot navigation based on openstreetmap geodata.
\newblock In {\em 13th International IEEE Conference on Intelligent Transportation Systems}, pages 1645--1650. IEEE, 2010.

\bibitem{osm_nav_2}
KAFA Samah, S~Ibrahim, N~Ghazali, M~Suffian, M~Mansor, and WA~Latif.
\newblock Mapping a hospital using openstreetmap and graphhopper: A navigation system.
\newblock {\em Bulletin of Electrical Engineering and Informatics}, 9(2):661--668, 2020.

\bibitem{osm_av1}
Julian Schmidt, Julian Jordan, Franz Gritschneder, Thomas Monninger, and Klaus Dietmayer.
\newblock Exploring navigation maps for learning-based motion prediction.
\newblock {\em arXiv preprint arXiv:2302.06195}, 2023.

\bibitem{liao2023osm}
Jing-Yan Liao, Parth Doshi, Zihan Zhang, David Paz, and Henrik Christensen.
\newblock Osm vs hd maps: Map representations for trajectory prediction.
\newblock {\em arXiv preprint arXiv:2311.02305}, 2023.

\bibitem{Zhou2021IROS_HDmap}
Yiyang Zhou, Yuichi Takeda, Masayoshi Tomizuka, and Wei Zhan.
\newblock Automatic construction of lane-level hd maps for urban scenes.
\newblock In {\em 2021 IEEE/RSJ International Conference on Intelligent Robots and Systems (IROS)}, pages 6649--6656, Prague, Czech Republic, 27 September - 01 October 2021.

\bibitem{road_statistics}
Federal Highway~Administration U.S. Department~of Transportation.
\newblock Highway statistics.
\newblock \url{http://www.fhwa.dot.gov/policyinformation/statistics.cfm}, last accessed on Feb 01, 2024.

\bibitem{Li_hdmapnet}
Qi~Li, Yue Wang, Yilun Wang, and Hang Zhao.
\newblock Hdmapnet: An online hd map construction and evaluation framework.
\newblock In {\em 2022 International Conference on Robotics and Automation (ICRA)}, pages 4628--4634, Philadelphia, PA, USA, 23-27 May 2022.

\bibitem{MapTR}
Bencheng Liao, Shaoyu Chen, Xinggang Wang, Tianheng Cheng, Qian Zhang, Wenyu Liu, and Chang Huang.
\newblock Maptr: Structured modeling and learning for online vectorized hd map construction.
\newblock In {\em International Conference on Learning Representations}, 2023.

\bibitem{maptrv2}
Bencheng Liao, Shaoyu Chen, Yunchi Zhang, Bo~Jiang, Qian Zhang, Wenyu Liu, Chang Huang, and Xinggang Wang.
\newblock Maptrv2: An end-to-end framework for online vectorized hd map construction.
\newblock {\em arXiv preprint arXiv:2308.05736}, 2023.

\bibitem{li2023toponet}
Tianyu Li, Li~Chen, Huijie Wang, Yang Li, Jiazhi Yang, Xiangwei Geng, Shengyin Jiang, Yuting Wang, Hang Xu, Chunjing Xu, Junchi Yan, Ping Luo, and Hongyang Li.
\newblock Graph-based topology reasoning for driving scenes.
\newblock {\em arXiv preprint arXiv:2304.05277}, 2023.

\bibitem{wu2023topomlp}
Dongming Wu, Jiahao Chang, Fan Jia, Yingfei Liu, Tiancai Wang, and Jianbing Shen.
\newblock Topomlp: A simple yet strong pipeline for driving topology reasoning.
\newblock {\em arXiv preprint arXiv:2310.06753}, 2023.

\bibitem{philion2020lift}
Jonah Philion and Sanja Fidler.
\newblock Lift, splat, shoot: Encoding images from arbitrary camera rigs by implicitly unprojecting to 3d.
\newblock {\em arXiv preprint arXiv:2008.05711}, 2020.

\bibitem{li2022bevformer}
Zhiqi Li, Wenhai Wang, Hongyang Li, Enze Xie, Chonghao Sima, Tong Lu, Yu~Qiao, and Jifeng Dai.
\newblock Bevformer: Learning bird’s-eye-view representation from multi-camera images via spatiotemporal transformers.
\newblock {\em arXiv preprint arXiv:2203.17270}, 2022.

\bibitem{liu2022bevfusion}
Zhijian Liu, Haotian Tang, Alexander Amini, Xinyu Yang, Huizi Mao, Daniela Rus, and Song Han.
\newblock Bevfusion: Multi-task multi-sensor fusion with unified bird's-eye view representation.
\newblock In {\em IEEE International Conference on Robotics and Automation (ICRA)}, ExCeL, London, UK, 29 May - 2 June 2023.

\bibitem{Sengupta12DenseVisual}
S.~{Sengupta}, P.~{Sturgess}, L.~{Ladický}, and P.~H.~S. {Torr}.
\newblock Automatic dense visual semantic mapping from street-level imagery.
\newblock In {\em 2012 IEEE/RSJ International Conference on Intelligent Robots and Systems}, pages 857--862, Vilamoura-Algarve, Portugal, 07-12 Oct 2012.

\bibitem{Sengupta13Stereo}
S.~{Sengupta}, E.~{Greveson}, A.~{Shahrokni}, and P.~H.~S. {Torr}.
\newblock Urban 3d semantic modelling using stereo vision.
\newblock In {\em 2013 IEEE International Conference on Robotics and Automation}, pages 580--585, Karlsruhe, Germany, 06-10 May 2013.

\bibitem{Paz2020IROSSemanticMapping}
David Paz, Hengyuan Zhang, Qinru Li, Hao Xiang, and Henrik~I. Christensen.
\newblock Probabilistic semantic mapping for urban autonomous driving applications.
\newblock In {\em 2020 IEEE/RSJ International Conference on Intelligent Robots and Systems (IROS)}, pages 2059--2064, Las Vegas, NV, USA, 24 October 2020.

\bibitem{liu2022vectormapnet}
Yicheng Liu, Tianyuan Yuan, Yue Wang, Yilun Wang, and Hang Zhao.
\newblock Vectormapnet: End-to-end vectorized hd map learning.
\newblock {\em arXiv preprint arXiv:2206.08920}, 2023.

\bibitem{AVL_semantic_mapping_sensors}
Hengyuan Zhang, Shashank Venkatramani, David Paz, Qinru Li, Hao Xiang, and Henrik~I. Christensen.
\newblock Probabilistic semantic mapping for autonomous driving in urban environments.
\newblock {\em Sensors}, 23(14), 2023.

\bibitem{tao2020hierarchicalMSHRNet}
Andrew Tao, Karan Sapra, and Bryan Catanzaro.
\newblock Hierarchical multi-scale attention for semantic segmentation.
\newblock {\em arXiv preprint arXiv:2005.10821}, 2020.

\bibitem{TensorRT}
NVIDIA.
\newblock Tensorrt open source software.
\newblock \url{https://github.com/NVIDIA/TensorRT}, last accessed on Feb 01, 2024.

\bibitem{Campos_orbslam3_2021}
Carlos Campos, Richard Elvira, Juan J.~Gómez Rodríguez, José~M. M.~Montiel, and Juan D.~Tardós.
\newblock Orb-slam3: An accurate open-source library for visual, visual–inertial, and multimap slam.
\newblock {\em IEEE Trans. Robot.}, 37(6):1874--1890, 2021.

\bibitem{Li2021liliom}
Kailai Li, Meng Li, and Uwe~D. Hanebeck.
\newblock Towards high-performance solid-state-lidar-inertial odometry and mapping.
\newblock {\em IEEE Robotics and Automation Letters}, 6(3):5167--5174, 2021.

\bibitem{He_2016_resnet}
Kaiming He, Xiangyu Zhang, Shaoqing Ren, and Jian Sun.
\newblock Deep residual learning for image recognition.
\newblock In {\em 2016 IEEE Conference on Computer Vision and Pattern Recognition (CVPR)}, pages 770--778, Las Vegas, NV, USA, 27-30 June 2016. IEEE.

\bibitem{Lin_2020_focal_loss}
Tsung-Yi Lin, Priya Goyal, Ross Girshick, Kaiming He, and Piotr Dollár.
\newblock Focal loss for dense object detection.
\newblock {\em IEEE Transactions on Pattern Analysis and Machine Intelligence}, 42(2):318--327, 2020.

\bibitem{MVD2017_Mapillary_Vistas}
Gerhard Neuhold, Tobias Ollmann, Samuel Rota~Bul\`o, and Peter Kontschieder.
\newblock The mapillary vistas dataset for semantic understanding of street scenes.
\newblock In {\em 2017 IEEE International Conference on Computer Vision (ICCV)}, pages 5000--5009, Venice, Italy, 22-29 October 2017.

\bibitem{christensen21:avl}
Henrik Christensen, David Paz, Hengyuan Zhang, Dominique Meyer, Hao Xiang, Yunhai Han, Yuhan Liu, Andrew Liang, Zheng Zhong, and Shiqi Tang.
\newblock Autonomous vehicles for micro-mobility.
\newblock {\em Auton. Intell. Syst.}, 1(11):1--35, Nov 2021.

\bibitem{esri}
Ersi.
\newblock World imagery.
\newblock \url{ https://www.arcgis.com/apps/mapviewer/index.html?layers=10df2279f9684e4a9f6a7f08febac2a9}, last accessed on Feb 01, 2024.

\end{thebibliography}

\end{document}